\documentclass[letterpaper]{article}  %Required
\usepackage{aaai19}  %Required
\usepackage{times}  %Required
\usepackage{helvet}  %Required
\usepackage{courier}  %Required
\usepackage{url}  %Required
\usepackage{graphicx}  %Required %%E
\usepackage{subcaption}
\usepackage{xcolor}
\usepackage[utf8]{inputenc}

\usepackage{comment}
\usepackage{enumitem}
\usepackage{multirow}
\usepackage{tabularx}

\usepackage{url}

\newcommand{\biburl}[1]{\url{#1}}

\usepackage{lipsum}

\setlist{noitemsep}

\newcommand{\citeauthoryearp}[1]{\citeauthor{#1} (\citeyear{#1})}

\title{Controllable Level Blending between Games using Variational Autoencoders}
\author{
Anurag Sarkar,\textsuperscript{1} Zhihan Yang\textsuperscript{2} and Seth Cooper\textsuperscript{1} \\
\textsuperscript{1}Northeastern University, Boston, Massachusetts, USA\\
\textsuperscript{2}Carleton College, Northfield, Minnesota, USA\\
sarkar.an@husky.neu.edu,
yangz2@carleton.edu,
se.cooper@northeastern.edu\\
}

\begin{document}

\maketitle

%================================================================================
% FIGURES

\newcommand{\XFIGUREDD}{
\begin{figure}[t]
\centering
\includegraphics[width=1\columnwidth]{figure/density_vs_difficulty}
\caption{\label{XFIGUREDD} Expressive range depicting Density vs Difficulty.}
\end{figure}
}

\newcommand{\XFIGUREall}{
\begin{figure*}[t!]
\centering
\includegraphics[width=0.675\columnwidth,trim={4cm 1cm 4cm 1cm},clip]{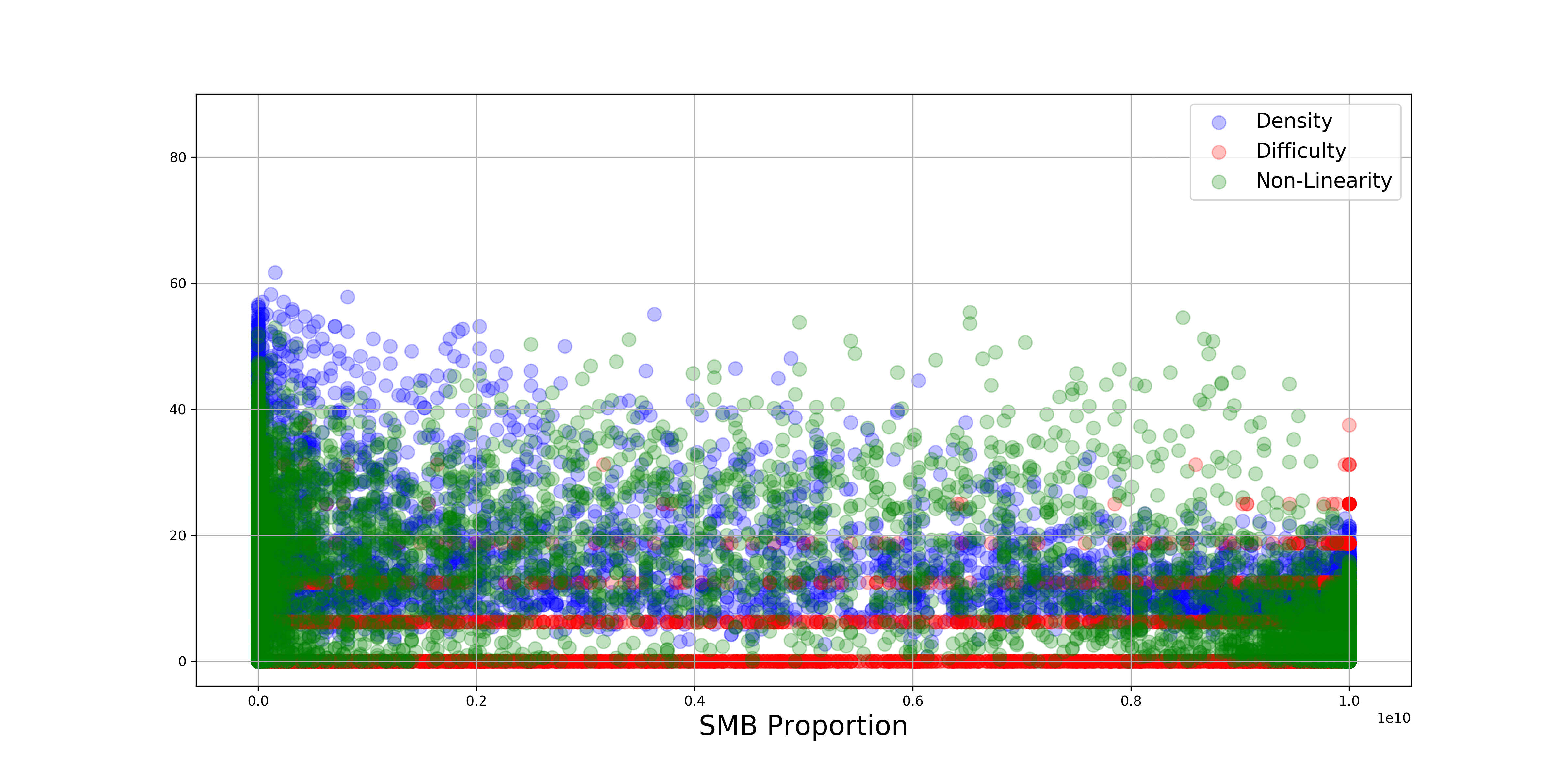}
\includegraphics[width=0.675\columnwidth,trim={4cm 1cm 4cm 1cm},clip]{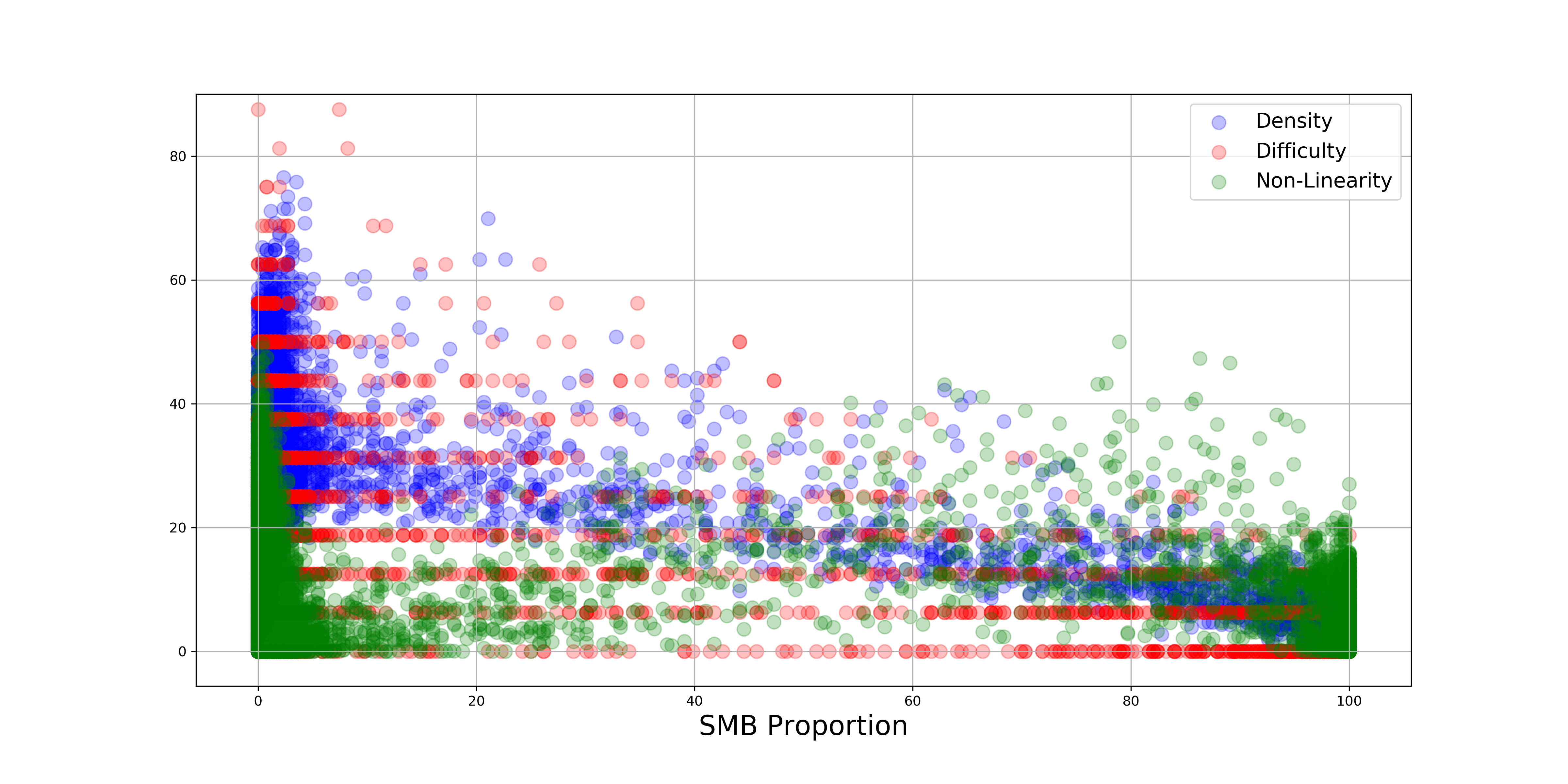}
\includegraphics[width=0.675\columnwidth,trim={4cm 1cm 4cm 1cm},clip]{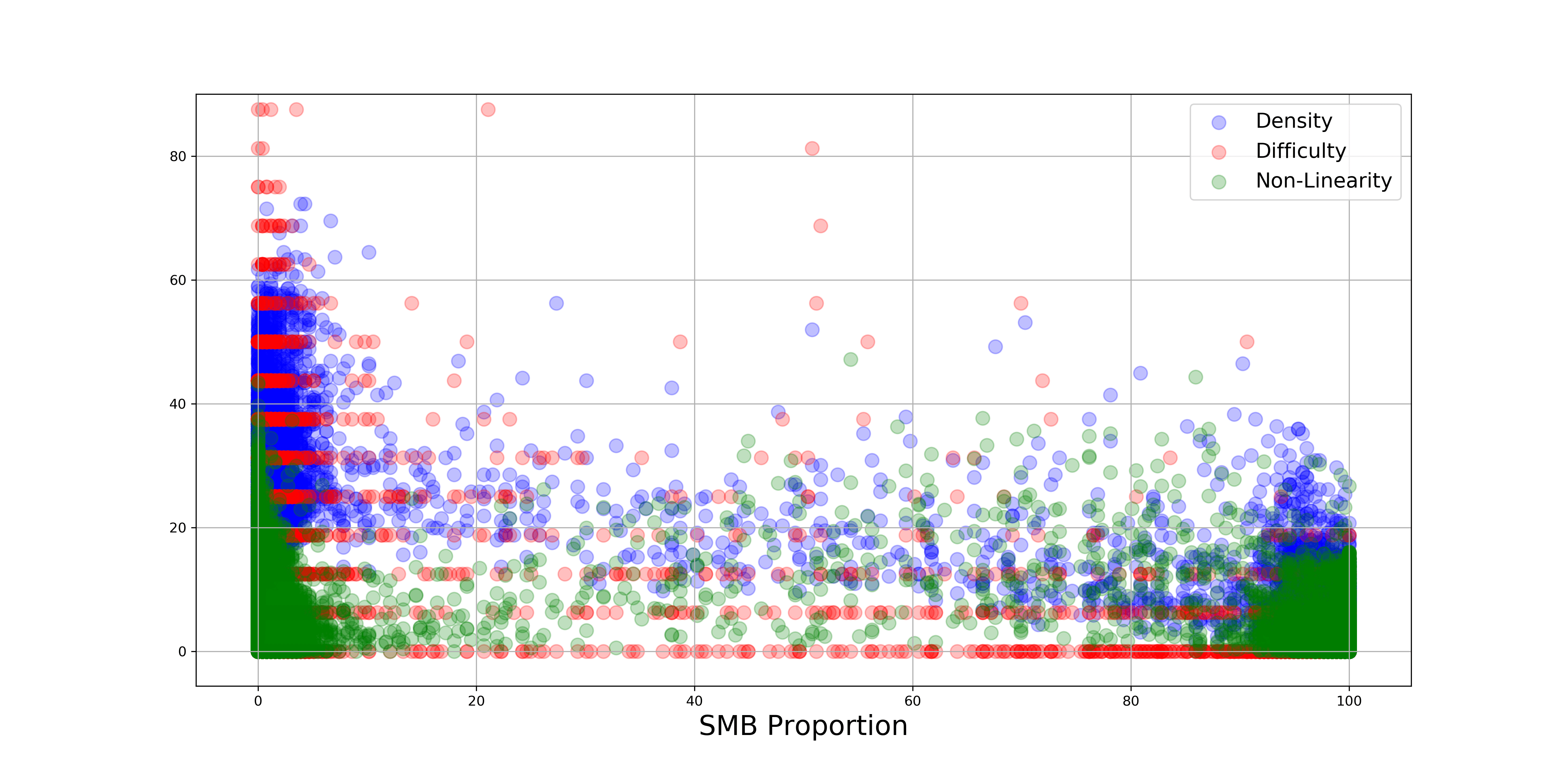}
\caption{\label{XFIGUREall} \textit{SMB Proportion} vs \textit{Density/Difficulty/Non-Linearity} for 10000 segments generated by VAE, GAN and VAE-GAN.}
\end{figure*}
}

\newcommand{\XFIGUREprops}{
\begin{figure*}[t]
\centering
\includegraphics[width=0.675\columnwidth]{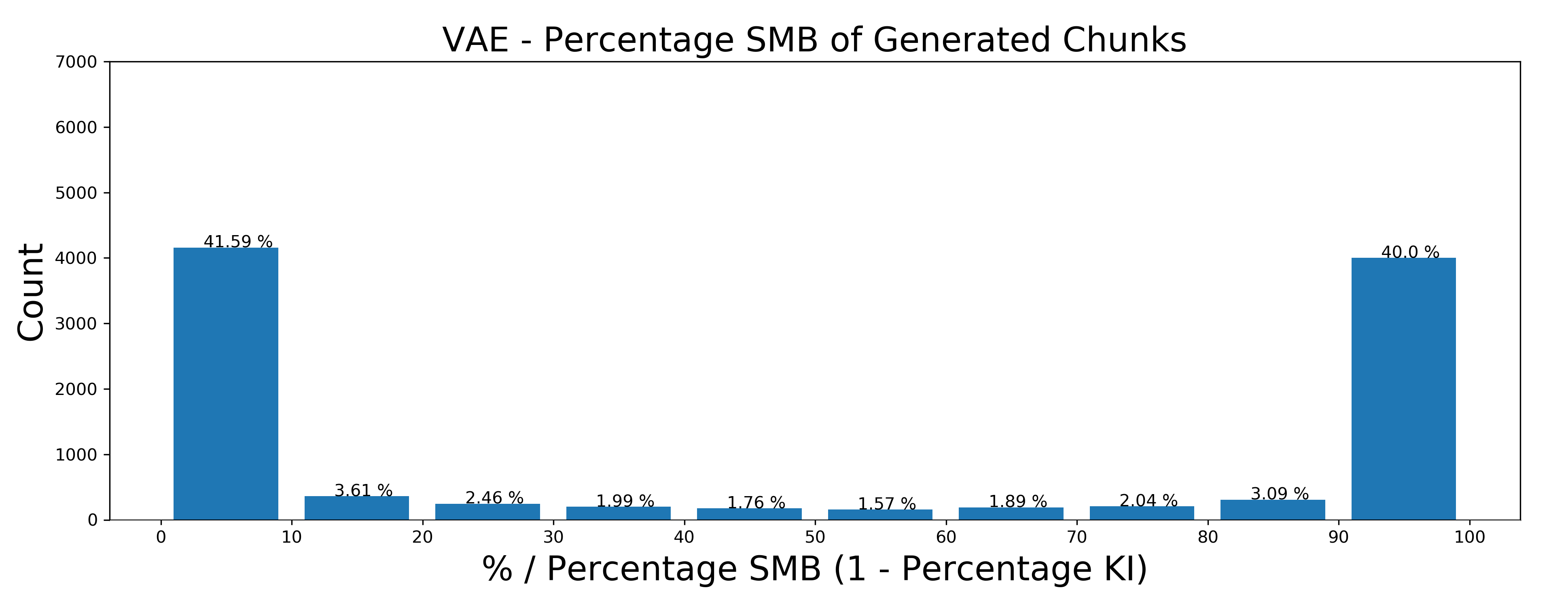}
\includegraphics[width=0.675\columnwidth]{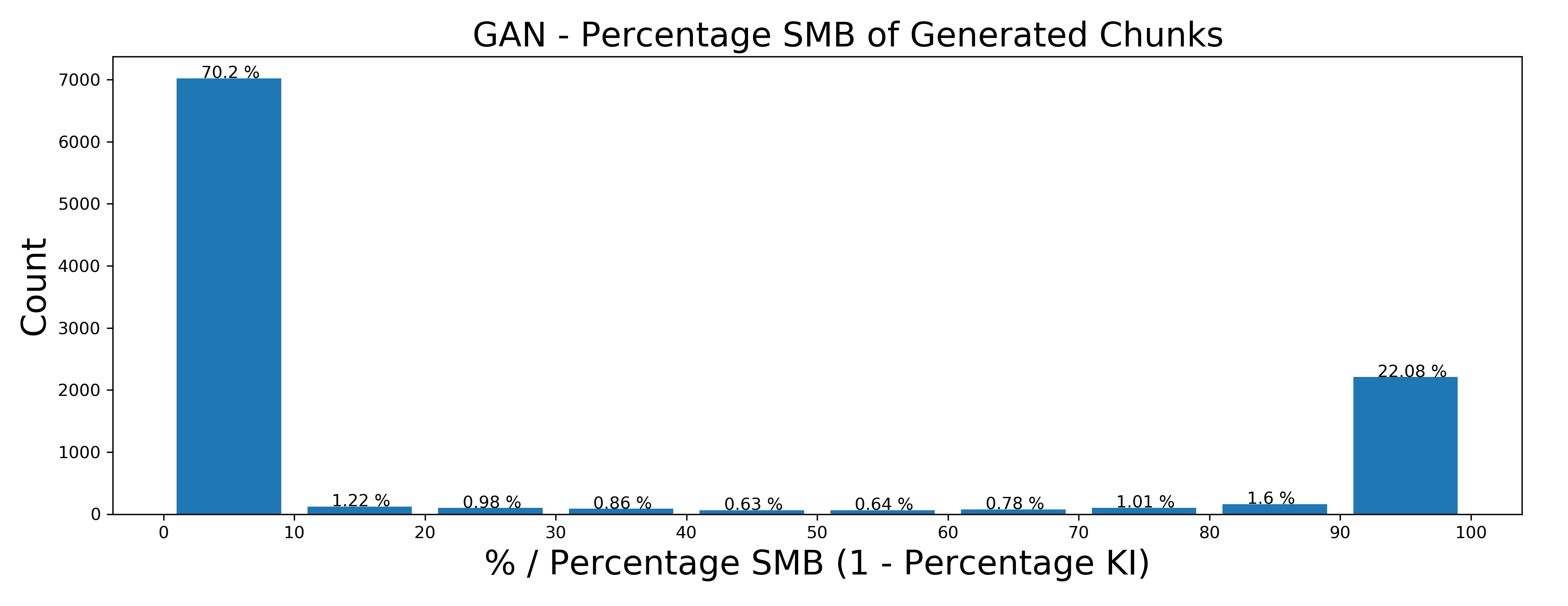} 
\includegraphics[width=0.675\columnwidth]{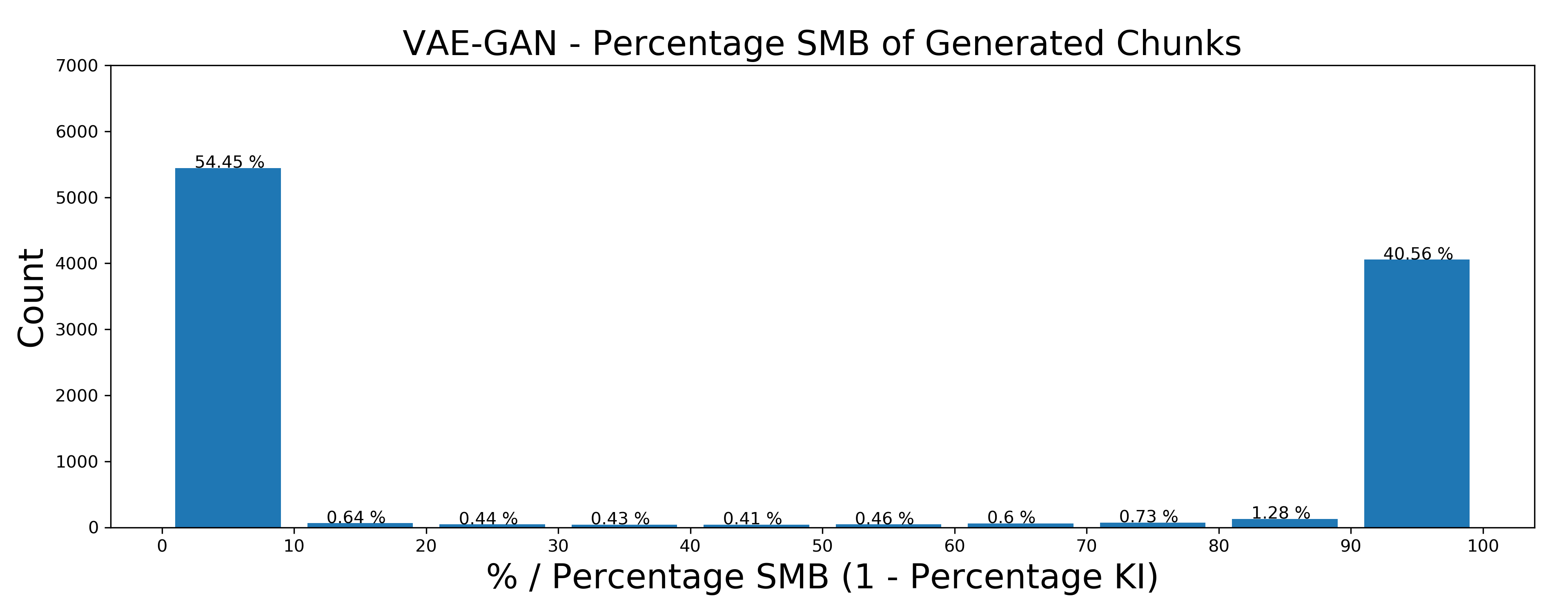}
\caption{\label{XFIGUREprops} Proportions of SMB and KI elements in segments generated by VAE, GAN and VAE-GAN respectively}
\end{figure*}
}

\newcommand{\XFIGURESK}{
\begin{figure}[t]
\centering
\includegraphics[width=1\columnwidth]{figure/proportions_back}
\caption{\label{XFIGURESK} Percentage of SMB vs KI tiles in 10000 randomly generated segments. These segments have good coverage of the space of percentages of both games i.e. the latent space captures level segments with all combinations of proportions of SMB and KI. }
\end{figure}
}

\newcommand{\XFIGUREevoacc}{
\begin{figure}[t]
\centering
\includegraphics[width=0.8\columnwidth]{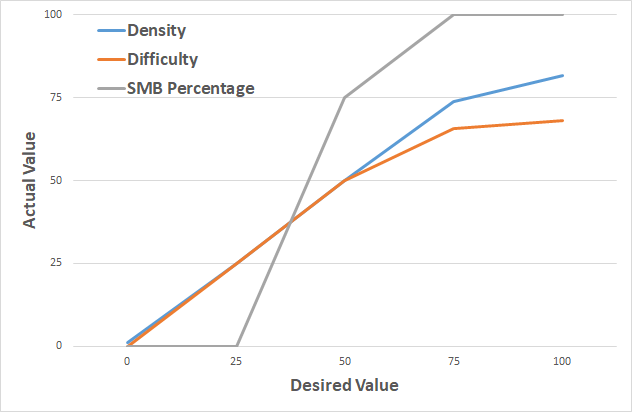}
\caption{\label{XFIGUREevoacc} Accuracy of evolving level segments optimizing density, difficulty and SMB Proportion. Values are the average of 100 evolved segments for each desired value.}
\end{figure}
}

\newcommand{\XFIGUREevo}{
\begin{figure*}[t]
\centering
\includegraphics[width=0.6\columnwidth]{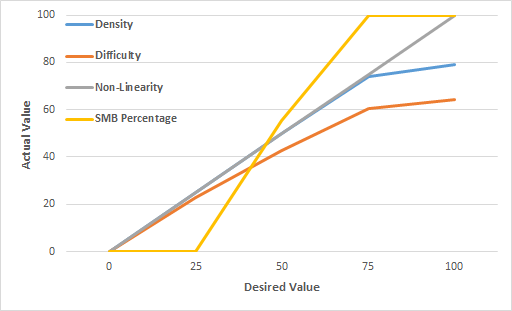}
\includegraphics[width=0.6\columnwidth]{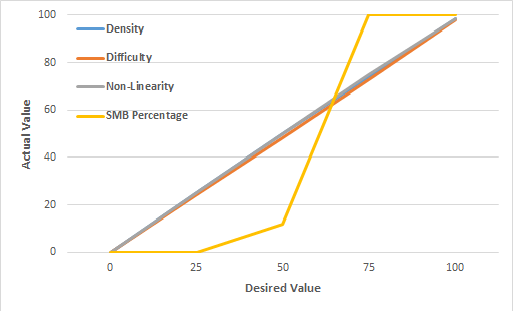}
\includegraphics[width=0.6\columnwidth]{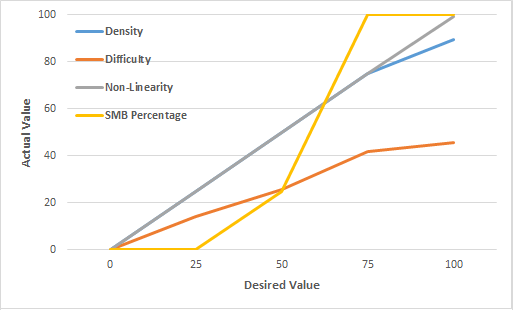}
\caption{\label{XFIGUREevo} Accuracy of evolving level segments optimizing \textit{Density}, \textit{Difficulty}, \textit{Non-Linearity} and \textit{SMB Proportion}. Values are the average of 100 evolved segments for each desired value. Results for VAE, GAN, VAE-GAN from left to right.}
\end{figure*}
}

%================================================================================
% TABLES

\newcolumntype{A}{>{\centering}p{1.25cm}}
\newcolumntype{B}{>{\centering\arraybackslash}p{1.25cm}}
\newcolumntype{C}{>{\centering}p{1.8cm}}
\newcolumntype{D}{>{\centering\arraybackslash}p{1.8cm}}
\newcolumntype{E}{>{\centering}p{2.9cm}}
\newcolumntype{F}{>{\centering\arraybackslash}p{2.9cm}}

\newcommand{\XTABLEencoding}{
\begin{table}
\begin{center}
\begin{tabular}{|l|l|l|l|}
\hline
\textit{Tile Type} & \textit{VGLC} & \textit{Integer} & \textit{Sprite} \\ \hline
SMB Ground & X & 0  & \includegraphics[width=10pt,height=10pt]{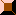}\\ \hline
SMB Breakable & S & 1 & \includegraphics[width=10pt,height=10pt]{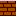}\\ \hline
SMB Background & - & 2 & \includegraphics[width=10pt,height=10pt]{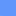}\\ \hline
SMB Full Question & ? & 3 & \includegraphics[width=10pt,height=10pt]{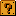}\\ \hline
SMB Empty Question & Q & 4 & \includegraphics[width=10pt,height=10pt]{figure/Q}\\ \hline
SMB Enemy & E & 5 & \includegraphics[width=10pt,height=10pt]{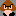}\\ \hline
SMB Pipe Top Left & $<$ & 6 & \includegraphics[width=10pt,height=10pt]{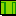} \\ \hline
SMB Pipe Top Right & $>$ & 7 & \includegraphics[width=10pt,height=10pt]{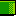} \\ \hline
SMB Pipe Bottom Left & [ & 8 & \includegraphics[width=10pt,height=10pt]{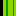} \\ \hline
SMB Pipe Bottom Right & ] & 9 & \includegraphics[width=10pt,height=10pt]{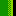} \\ \hline
SMB Coin & o & 10 &
\includegraphics[width=10pt,height=10pt]{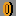} \\ \hline
KI Platform & T & 11 & \includegraphics[width=10pt,height=10pt]{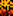} \\ \hline
KI Movable Platform & M & 12 & \includegraphics[width=10pt,height=10pt]{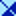} \\ \hline
KI Door & D & 13 & \includegraphics[width=10pt,height=10pt]{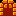} \\ \hline
KI Ground & \# & 14 & \includegraphics[width=10pt,height=10pt]{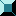} \\ \hline
KI Hazard & H & 15 & \includegraphics[width=10pt,height=10pt]{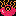} \\ \hline
KI Background & - & 16 & \includegraphics[width=10pt,height=10pt]{figure/back} \\ \hline
\end{tabular}
\caption{\label{XTABLEencoding} Encodings used for level representation}
\end{center}
\end{table}
}

\newcommand{\XFIGUREint}{
\begin{figure}[t]
\centering
\setlength\tabcolsep{1pt}
\begin{tabular}{cccccc}
\includegraphics[width=0.0725\textwidth]{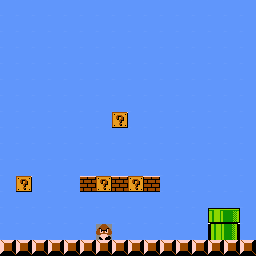}
\includegraphics[width=0.0725\textwidth]{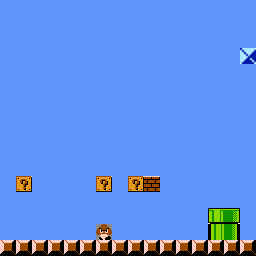}
\includegraphics[width=0.0725\textwidth]{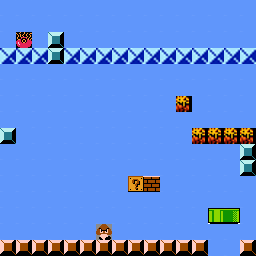}
\includegraphics[width=0.0725\textwidth]{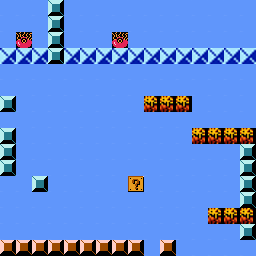}
\includegraphics[width=0.0725\textwidth]{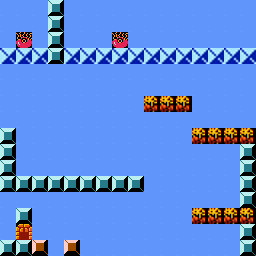}
\includegraphics[width=0.0725\textwidth]{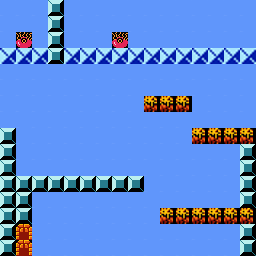}
\end{tabular}
\caption{\label{XFIGUREint} Transitioning from SMB (left) to KI (right) by interpolating between corresponding latent vectors.}
\end{figure}
}

\newcommand{\XFIGUREsint}{
\begin{figure}[t]
\centering
\setlength\tabcolsep{1pt}
\begin{tabular}{ccccc}
\includegraphics[width=0.0725\textwidth]{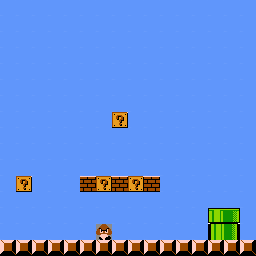}
\includegraphics[width=0.0725\textwidth]{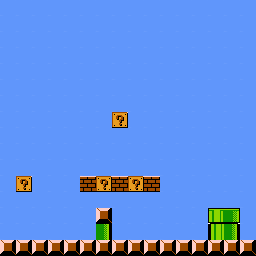}
\includegraphics[width=0.0725\textwidth]{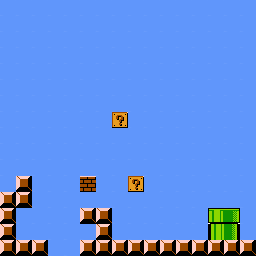}
\includegraphics[width=0.0725\textwidth]{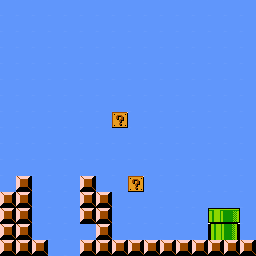}
\includegraphics[width=0.0725\textwidth]{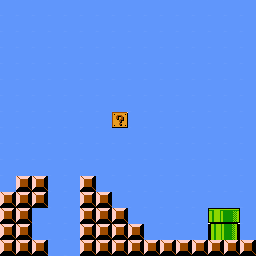}
\includegraphics[width=0.0725\textwidth]{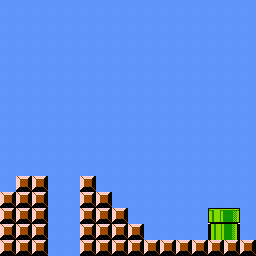}
\end{tabular}
\caption{\label{XFIGUREsint} Interpolating between segments of Mario 1-1 generates those not in the actual level as in the middle four.}
\end{figure}
}

\newcommand{\XFIGUREmaxtt}{
\begin{figure}[t]
\centering
\setlength\tabcolsep{1pt}
\begin{tabular}{cccccccc}
\rotatebox{90}{\tiny{\hspace{12pt}\textbf{VAE}}} &
\includegraphics[width=0.07\textwidth]{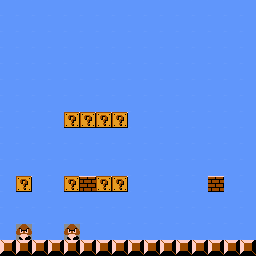} &
\includegraphics[width=0.07\textwidth]{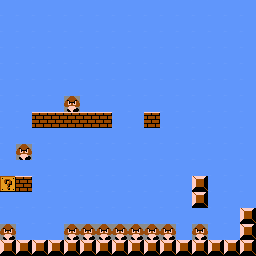} &
\includegraphics[width=0.07\textwidth]{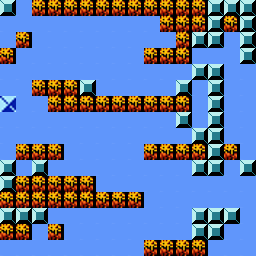} &
\includegraphics[width=0.07\textwidth]{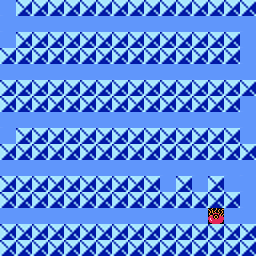} &
\includegraphics[width=0.07\textwidth]{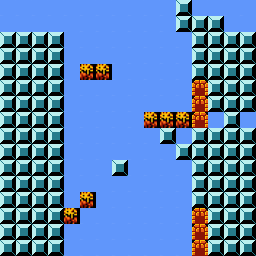} &
\includegraphics[width=0.07\textwidth]{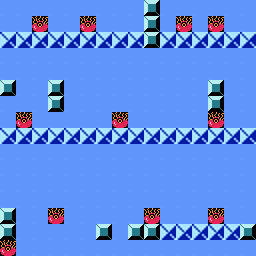} \\
\rotatebox{90}{\tiny{\hspace{12pt}\textbf{GAN}}} &
\includegraphics[width=0.07\textwidth]{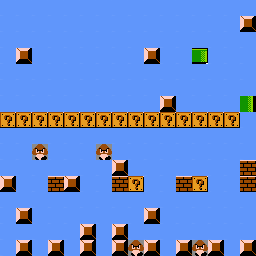} &
\includegraphics[width=0.07\textwidth]{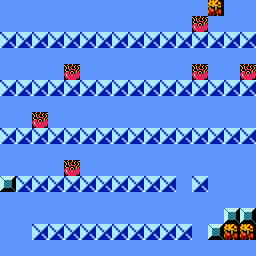} &
\includegraphics[width=0.07\textwidth]{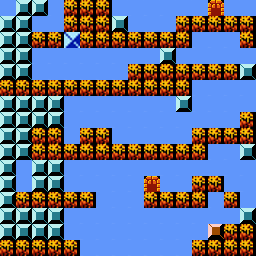} &
\includegraphics[width=0.07\textwidth]{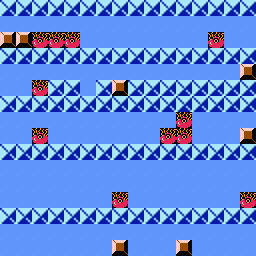} &
\includegraphics[width=0.07\textwidth]{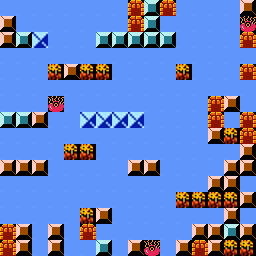} &
\includegraphics[width=0.07\textwidth]{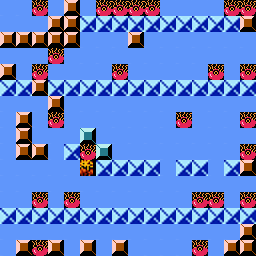} \\
\rotatebox{90}{\tiny{\hspace{3pt}\textbf{VAE-GAN}}} &
\includegraphics[width=0.07\textwidth]{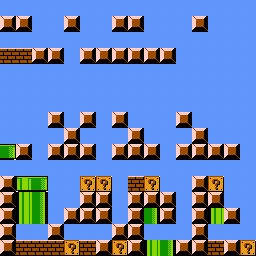} &
\includegraphics[width=0.07\textwidth]{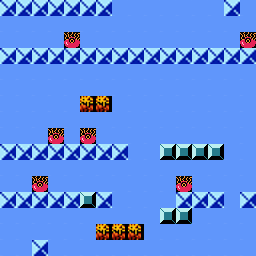} &
\includegraphics[width=0.07\textwidth]{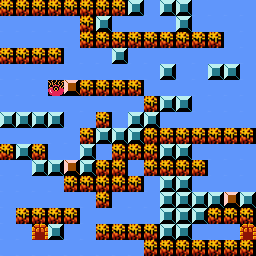} &
\includegraphics[width=0.07\textwidth]{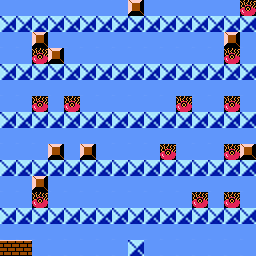} &
\includegraphics[width=0.07\textwidth]{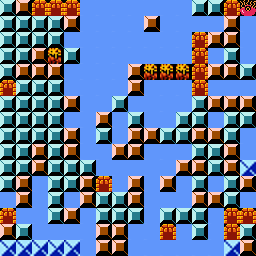} &
\includegraphics[width=0.07\textwidth]{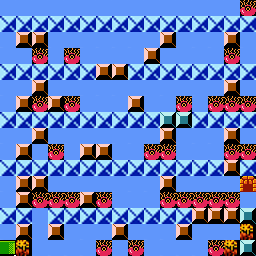} \\
& $?$ & E & T & M & D & H
\end{tabular}
\caption{\label{XFIGUREmaxtt} Evolved segments maximizing given tile type.}
\end{figure}
}

\newcommand{\XFIGUREcorners}{
\begin{figure*}[t]
\centering
\setlength\tabcolsep{2pt}
\begin{tabular}{ccc}
\includegraphics[width=0.25\textwidth]{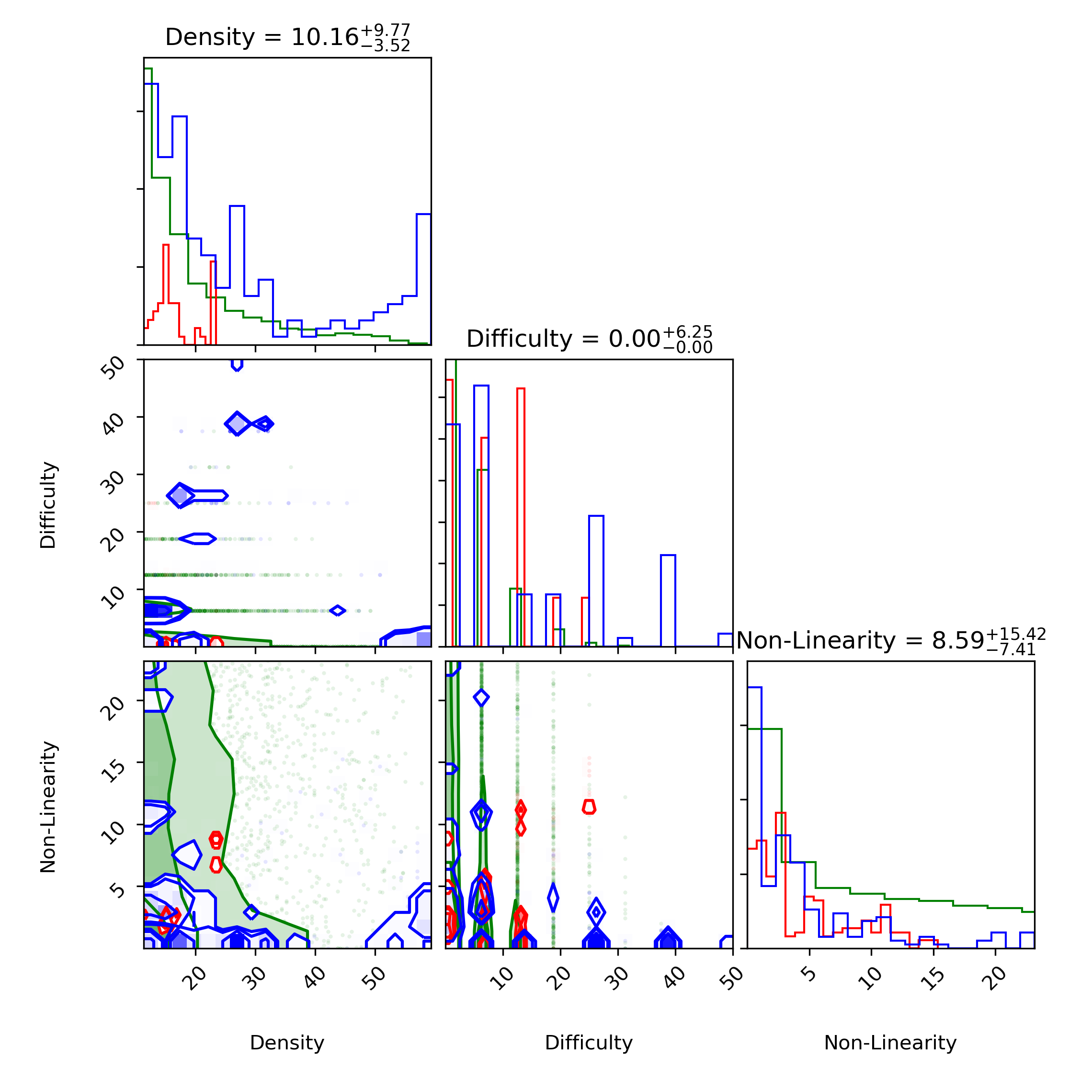}
\includegraphics[width=0.25\textwidth]{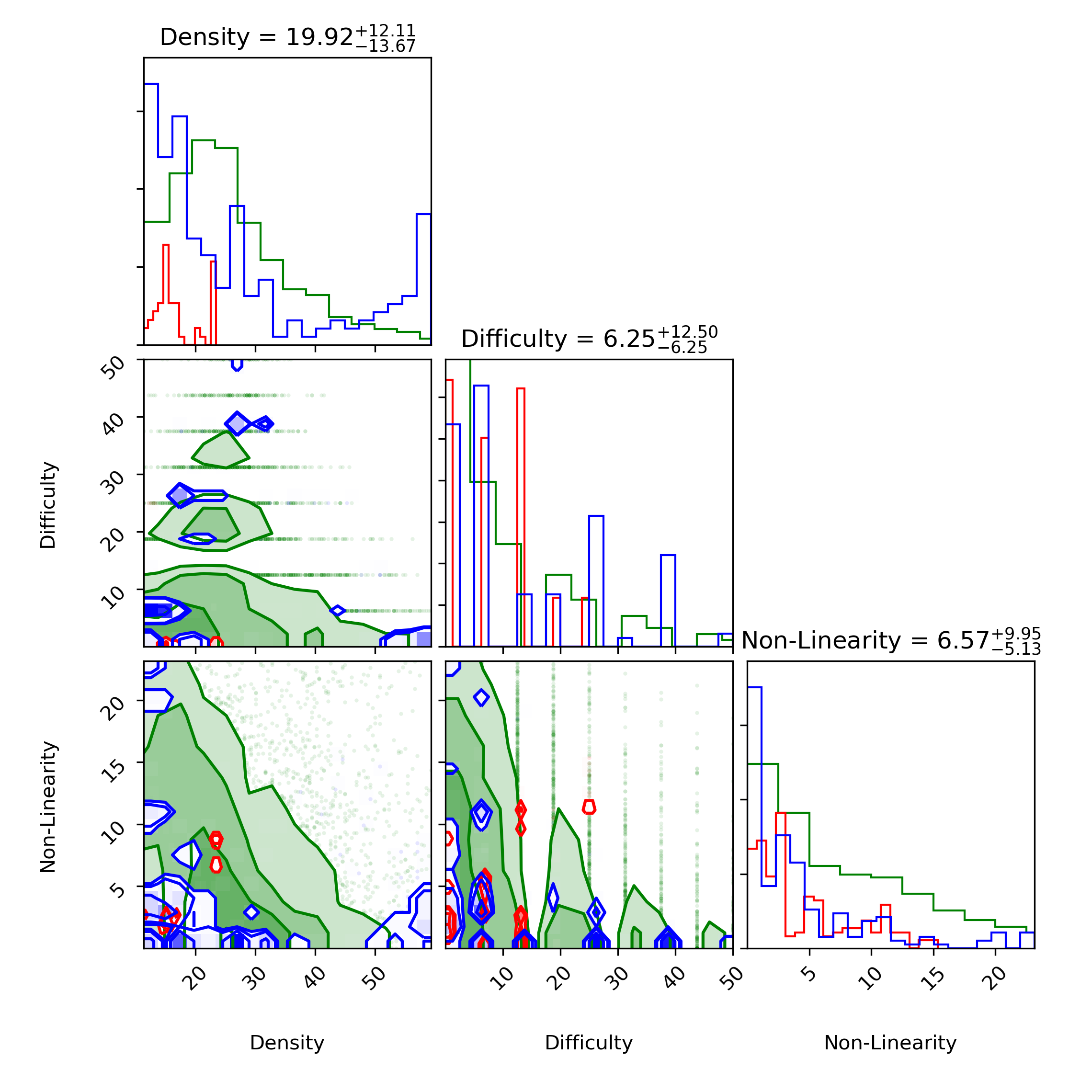}
\includegraphics[width=0.25\textwidth]{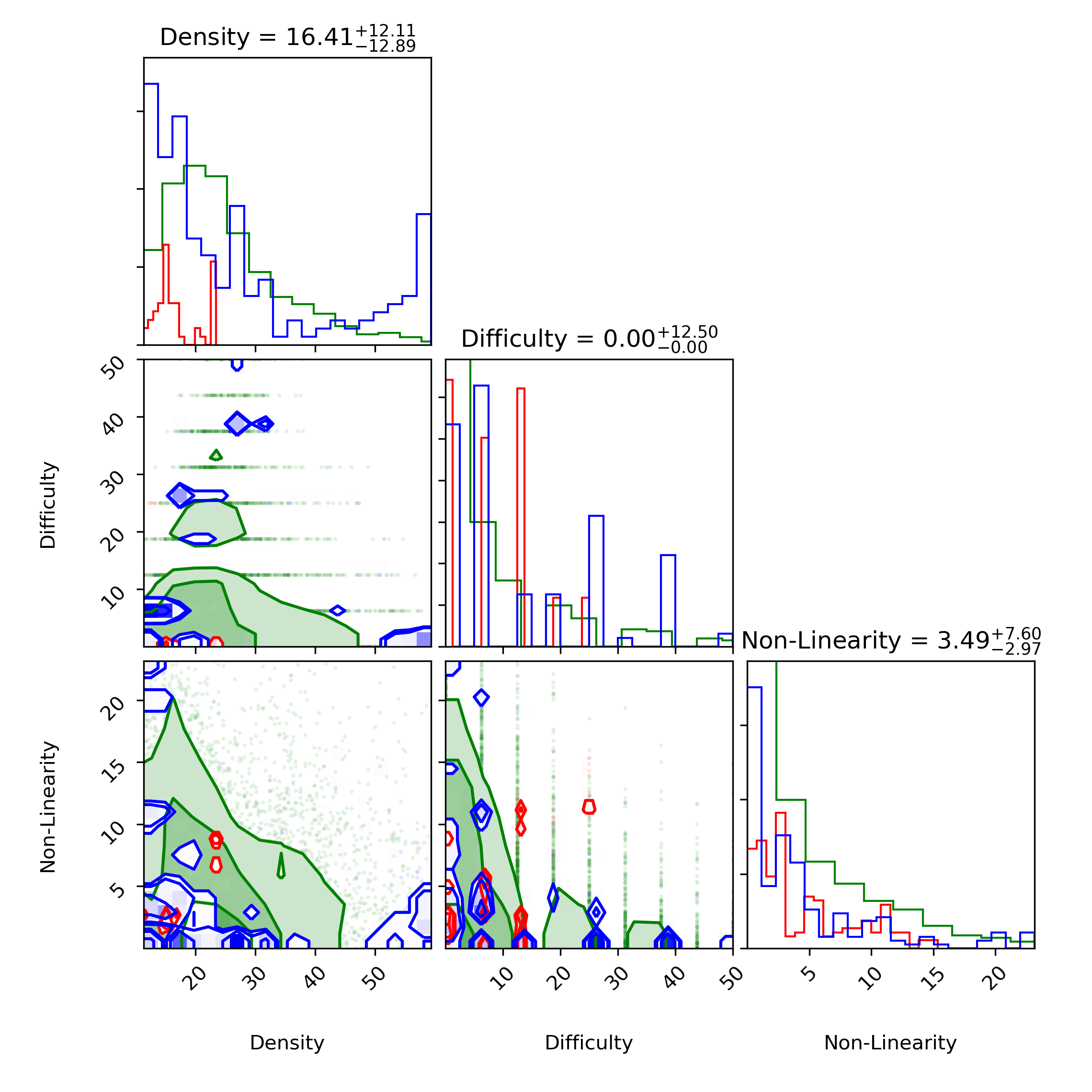}
\end{tabular}
\caption{\label{XFIGUREcorners} Corner plots for VAE, GAN and VAE-GAN respectively depicting generated (green), SMB (red) and KI (blue) levels.}
\end{figure*}
}

\newcommand{\XFIGUREdensity}{
\begin{figure}[t]
\centering
\setlength\tabcolsep{2pt}
\begin{tabular}{cccccc}
\rotatebox{90}{\scriptsize{\hspace{12pt}\textbf{VAE}}} &
\includegraphics[width=0.08\textwidth]{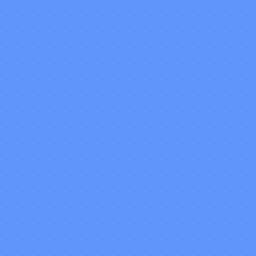} &
\includegraphics[width=0.08\textwidth]{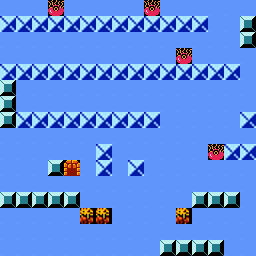} &
\includegraphics[width=0.08\textwidth]{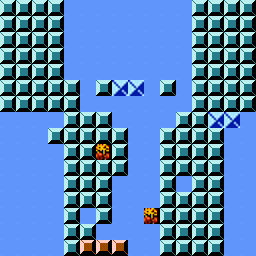} &
\includegraphics[width=0.08\textwidth]{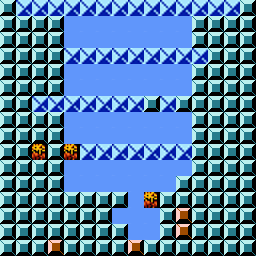} &
\includegraphics[width=0.08\textwidth]{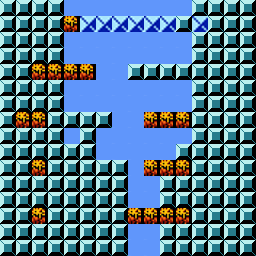} \\
\rotatebox{90}{\scriptsize{\hspace{12pt}\textbf{GAN}}} &
\includegraphics[width=0.08\textwidth]{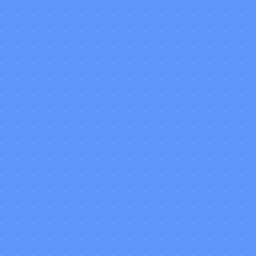} &
\includegraphics[width=0.08\textwidth]{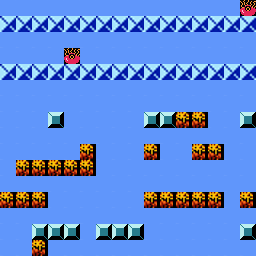} &
\includegraphics[width=0.08\textwidth]{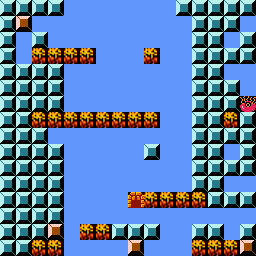} &
\includegraphics[width=0.08\textwidth]{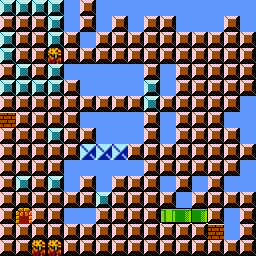} &
\includegraphics[width=0.08\textwidth]{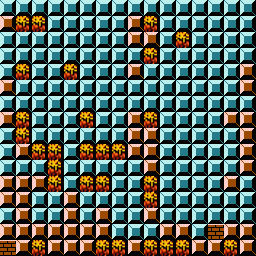} \\
\rotatebox{90}{\scriptsize{\hspace{3pt}\textbf{VAE-GAN}}} &
\includegraphics[width=0.08\textwidth]{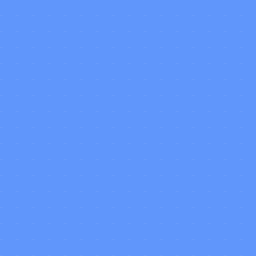} &
\includegraphics[width=0.08\textwidth]{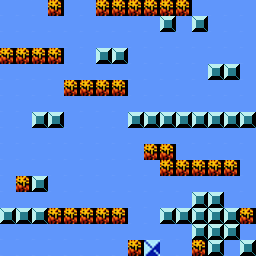} &
\includegraphics[width=0.08\textwidth]{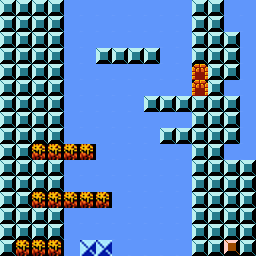} &
\includegraphics[width=0.08\textwidth]{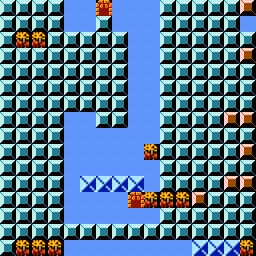} &
\includegraphics[width=0.08\textwidth]{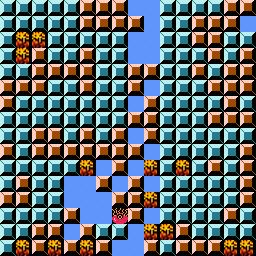} \\
 & 0\% & 25\% & 50\% & 75\% & 100\%
\end{tabular}
\caption{\label{XFIGUREdensity} Evolved segments optimizing for \textit{Density}.}
\end{figure}
}

\newcommand{\XFIGUREdifficulty}{
\begin{figure}[t]
\centering
\setlength\tabcolsep{2pt}
\begin{tabular}{cccccc}
\rotatebox{90}{\scriptsize{\hspace{12pt}\textbf{VAE}}} &
\includegraphics[width=0.08\textwidth]{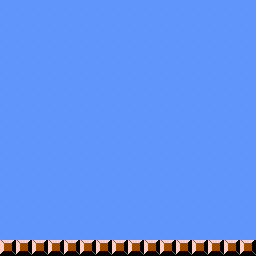} &
\includegraphics[width=0.08\textwidth]{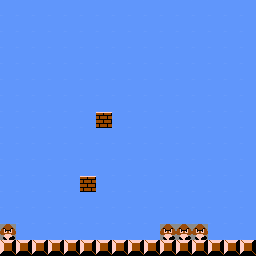} &
\includegraphics[width=0.08\textwidth]{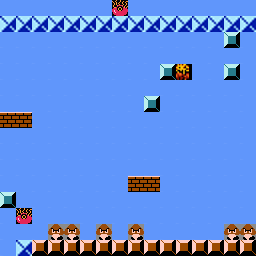} &
\includegraphics[width=0.08\textwidth]{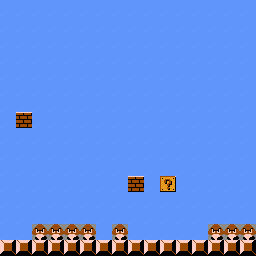} &
\includegraphics[width=0.08\textwidth]{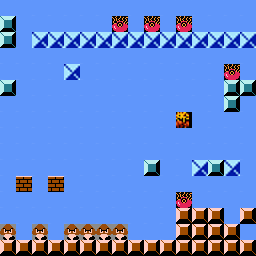} \\
\rotatebox{90}{\scriptsize{\hspace{12pt}\textbf{GAN}}} &
\includegraphics[width=0.08\textwidth]{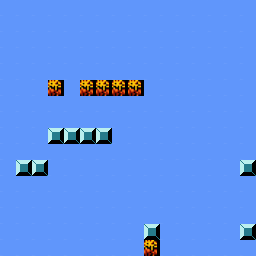} &
\includegraphics[width=0.08\textwidth]{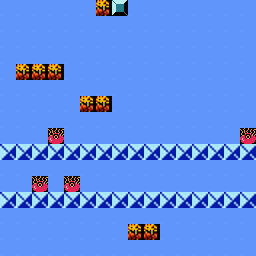} &
\includegraphics[width=0.08\textwidth]{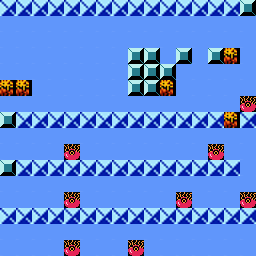} &
\includegraphics[width=0.08\textwidth]{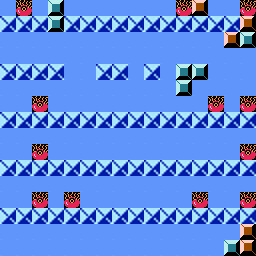} &
\includegraphics[width=0.08\textwidth]{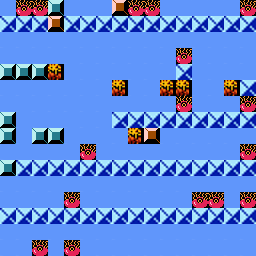} \\
\rotatebox{90}{\scriptsize{\hspace{3pt}\textbf{VAE-GAN}}} &
\includegraphics[width=0.08\textwidth]{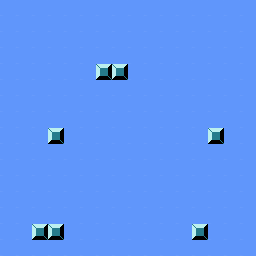} &
\includegraphics[width=0.08\textwidth]{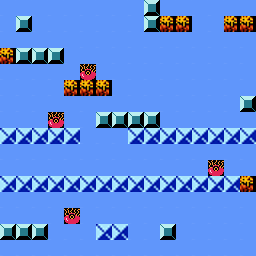} &
\includegraphics[width=0.08\textwidth]{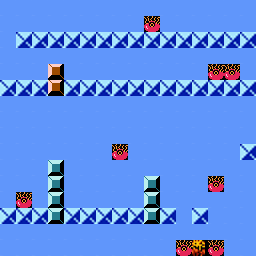} &
\includegraphics[width=0.08\textwidth]{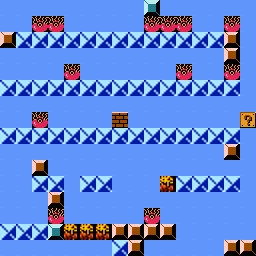} &
\includegraphics[width=0.08\textwidth]{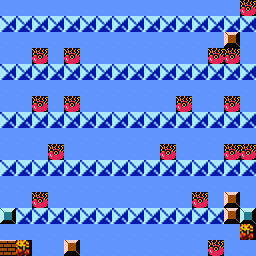} \\
& 0\% & 25\% & 50\% & 75\% & 100\%
\end{tabular}
\caption{\label{XFIGUREdifficulty} Evolved segments optimizing for \textit{Difficulty}.}
\end{figure}
}

\newcommand{\XFIGUREnonlinearity}{
\begin{figure}[t]
\centering
\setlength\tabcolsep{2pt}
\begin{tabular}{cccccc}
\rotatebox{90}{\scriptsize{\hspace{12pt}\textbf{VAE}}} &
\includegraphics[width=0.08\textwidth]{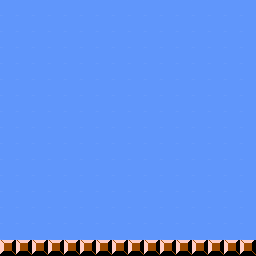} &
\includegraphics[width=0.08\textwidth]{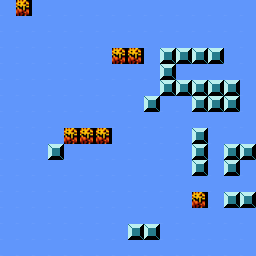} &
\includegraphics[width=0.08\textwidth]{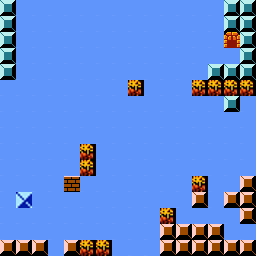} &
\includegraphics[width=0.08\textwidth]{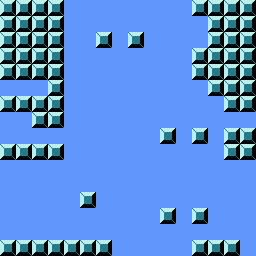} &
\includegraphics[width=0.08\textwidth]{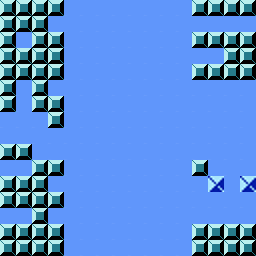} \\
\rotatebox{90}{\scriptsize{\hspace{12pt}\textbf{GAN}}} &
\includegraphics[width=0.08\textwidth]{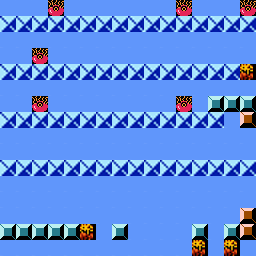} &
\includegraphics[width=0.08\textwidth]{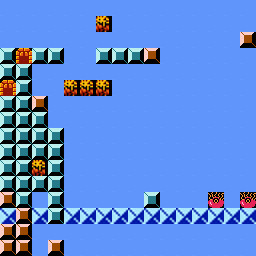} &
\includegraphics[width=0.08\textwidth]{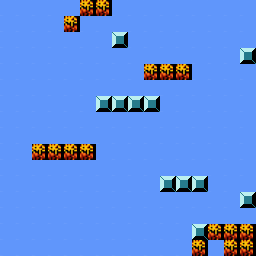} &
\includegraphics[width=0.08\textwidth]{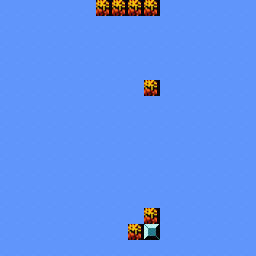} &
\includegraphics[width=0.08\textwidth]{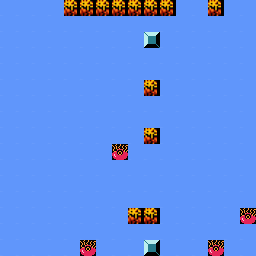} \\
\rotatebox{90}{\scriptsize{\hspace{3pt}\textbf{VAE-GAN}}} &
\includegraphics[width=0.08\textwidth]{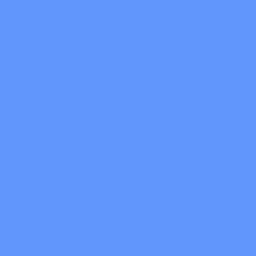} &
\includegraphics[width=0.08\textwidth]{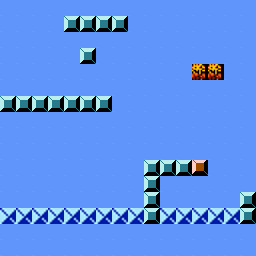} &
\includegraphics[width=0.08\textwidth]{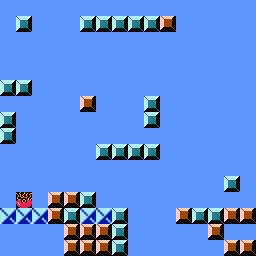} &
\includegraphics[width=0.08\textwidth]{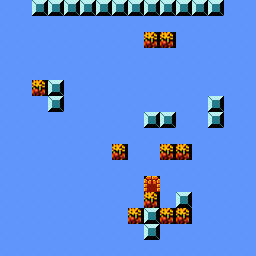} &
\includegraphics[width=0.08\textwidth]{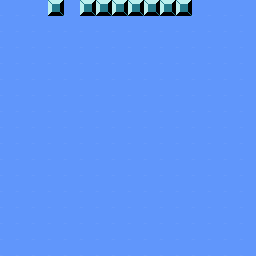} \\
& 0\% & 25\% & 50\% & 75\% & 100\%
\end{tabular}
\caption{\label{XFIGUREnonlinearity} Evolved segments optimizing for \textit{Non-Linearity}.}
\end{figure}
}

\newcommand{\XFIGUREblending}{
\begin{figure}[t]
\centering
\setlength\tabcolsep{2pt}
\begin{tabular}{cccccc}
\rotatebox{90}{\scriptsize{\hspace{12pt}\textbf{VAE}}} &
\includegraphics[width=0.08\textwidth]{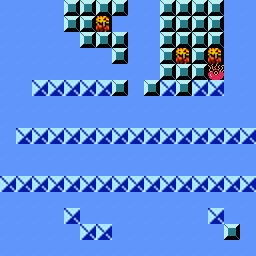} &
\includegraphics[width=0.08\textwidth]{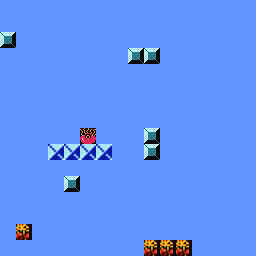} &
\includegraphics[width=0.08\textwidth]{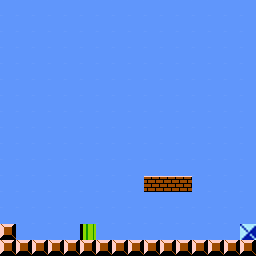} &
\includegraphics[width=0.08\textwidth]{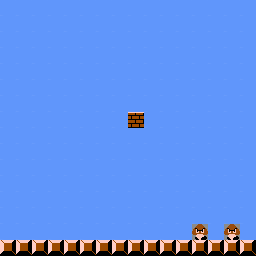} &
\includegraphics[width=0.08\textwidth]{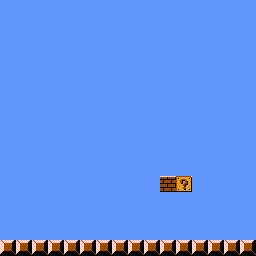} \\
\rotatebox{90}{\scriptsize{\hspace{12pt}\textbf{GAN}}} &
\includegraphics[width=0.08\textwidth]{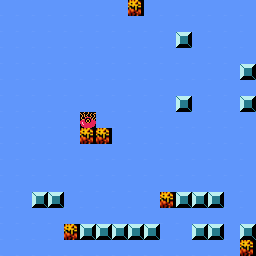} &
\includegraphics[width=0.08\textwidth]{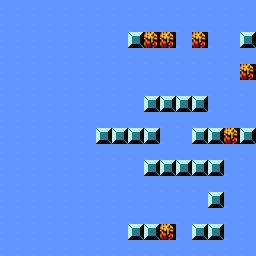} &
\includegraphics[width=0.08\textwidth]{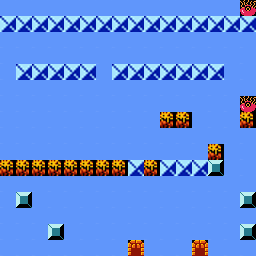} &
\includegraphics[width=0.08\textwidth]{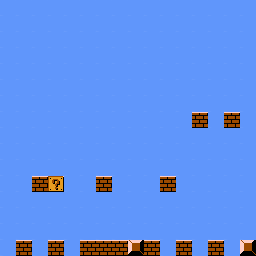} &
\includegraphics[width=0.08\textwidth]{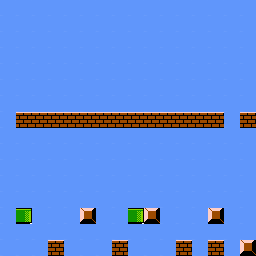} \\
\rotatebox{90}{\scriptsize{\hspace{3pt}\textbf{VAE-GAN}}} &
\includegraphics[width=0.08\textwidth]{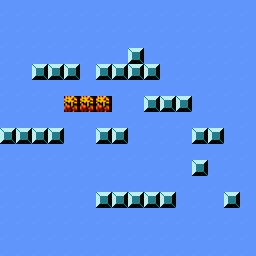} &
\includegraphics[width=0.08\textwidth]{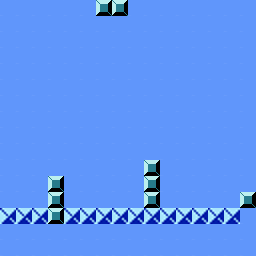} &
\includegraphics[width=0.08\textwidth]{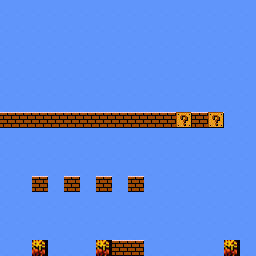} &
\includegraphics[width=0.08\textwidth]{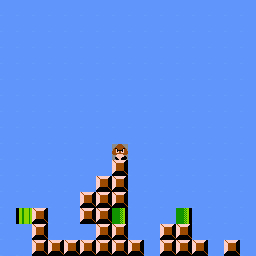} &
\includegraphics[width=0.08\textwidth]{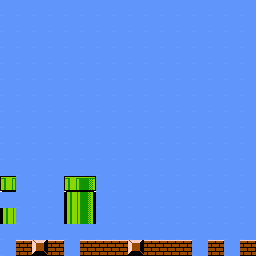} \\
& 0\% & 25\% & 50\% & 75\% & 100\%
\end{tabular}
\caption{\label{XFIGUREblending} Evolved segments optimizing \textit{SMB Proportion}.} 
\end{figure}
}

%================================================================================
\begin{abstract}
Previous work explored blending levels from existing games to create levels for a new game that mixes properties of the original games. In this paper, we use Variational Autoencoders (VAEs) for improving upon such techniques. VAEs are artificial neural networks that learn and use latent representations of datasets to generate novel outputs. We train a VAE on level data from \textit{Super Mario Bros.} and \textit{Kid Icarus}, enabling it to capture the latent space spanning both games. We then use this space to generate level segments that combine properties of levels from both games. Moreover, by applying evolutionary search in the latent space, we evolve level segments satisfying specific constraints. We argue that these affordances make the VAE-based approach especially suitable for co-creative level design and compare its performance with similar generative models like the GAN and the VAE-GAN.
\end{abstract}

%================================================================================
\section{Introduction}
Procedural content generation (PCG) refers to automated generation of content for games and has been a popular area of research \cite{shaker_procedural_2016} with methods for PCG using a variety of techniques, most commonly search \cite{togelius_search-based_2011}, grammars \cite{smith_tanagra_2011} and constraint solving \cite{smith_answer_2011}. A newly emerging subfield of PCG research is PCGML or PCG via Machine Learning \cite{summerville_procedural_2017}, referring to procedurally generating content via models trained on existing game data using machine learning. This allows models to capture properties of games that designers may want to emulate while also helping alleviate the burden of hand-crafting rules and reducing designer bias.

Recently, \citeauthoryearp{sarkar_blending_2018} showed that LSTMs trained on levels from \textit{Super Mario Bros.} and \textit{Kid Icarus} can generate levels whose properties are a mix of those of levels from the original games. This work leveraged past successes in ML-based level generation \cite{summerville_learning_2015} and blending \cite{guzdial_learning_2016} to move towards implementing a co-creative system similar to the VGDL-based game blending framework proposed by \citeauthoryearp{gow_towards_2015} wherein a novel game is generated by combining the mechanics and aesthetics of two existing games. While successful in demonstrating the feasibility of generating blended levels using models trained on data from separate games, the LSTM approach did blending by taking turns in generating segments of the two games. This allows generation of blended levels but not more fine-grained blending of elements from different games within a level segment. Also, though it lets designers specify the proportion of each game to blend in the levels, it does not let them control different level properties in the blending and generation process.

To address these issues, we use Variational Autoencoders (VAEs) to blend levels from different games and show that by capturing the latent design space across both games, the VAE enables more holistic blending of level properties, allows for the generation of level segments that optimize certain functions and satisfy specific properties, and is more conducive to co-creative level design. Specifically, we train a VAE on one level each from \textit{Super Mario Bros.} and \textit{Kid Icarus} which thus learns a latent representation spanning both games. Using evolutionary search in this space lets us evolve level segments satisfying different designer-specified constraints as well as the proportions of \textit{Super Mario Bros.} and \textit{Kid Icarus} elements present within them. Due to the nature of the VAE, designers can also specify the segments they want to blend or generate variations of without having to use the latent representation as in past related work using Generative Adversarial Networks (GANs) \cite{volz_evolving_2018}.

This work contributes an exploratory study on using VAEs to blend levels from different games. We present an evaluation of the approach that suggests that VAEs are better suited to blending levels across games than GANs and conclude with a discussion on how VAEs can inform co-creative level design in the future.

%================================================================================
\section{Background and Related Work}

\subsubsection{PCG via Machine Learning} 
Recent years have seen significant research on 2D level generation using machine learning. N-gram models \cite{dahlskog_linear_2014}, Markov models \cite{snodgrass_experiments_2014}, PGMs \cite{guzdial_game_2016} and LSTMs \cite{summerville_super_2017} have all been used to generate levels for \textit{Super Mario Bros.} while Bayes Nets have been used for generating dungeons for \textit{The Legend of Zelda} \cite{summerville_sampling_2015}. Similar to our work, \citeauthoryearp{snodgrass_approach_2016} use levels from multiple games, including the two that we used, but differ in using Markov chains rather than VAEs and in focusing on translating levels from one game to another rather than blending levels. Most closely related to our work is that of \citeauthoryearp{jain_autoencoders_2016} and \citeauthoryearp{volz_evolving_2018} in using autoencoders and GANs respectively to create \textit{Super Mario Bros.} levels. The latter also applied Covariance Matrix Adaptation to evolve levels with certain properties within the GAN's latent space. We similarly train a VAE and evolve levels in its latent space but unlike GANs, which only accept latent space vectors as inputs, VAEs can accept as inputs both level segments as well as latent vectors, thus offering more control over generation.

\subsubsection{Level and Game Blending}
Past work has focused on blending levels from the same game, specifically \textit{Super Mario Bros.}, by applying principles of conceptual blending \cite{fauconnier_conceptual_1998}. \citeauthoryearp{guzdial_learning_2016} blended different Mario level generation models while also exploring different blending strategies for generating levels \cite{guzdial_combinatorial_2017}. Beyond levels, \citeauthoryearp{gow_towards_2015} proposed a framework for blending entire games. They created a new game \textit{Frolda} by combining the VGDL \cite{schaul_extensible_2014} specifications of \textit{Frogger} and \textit{Zelda}. To move towards using ML-based level blending to implement a similar framework but not restricted to VGDL, \citeauthoryearp{sarkar_blending_2018} used LSTMs to blend together levels from \textit{Super Mario Bros.} and \textit{Kid Icarus}. In this paper, we build on this latter work by replacing LSTMs with VAEs. This enables more holistic level blending and the ability to use the VAE's learned latent space to interpolate between levels and evolve new ones based on various properties. Similar to blending games, recent work by \citeauthoryearp{guzdial_automated_2018} looked into generating new games by recombining existing ones using a process termed \textit{conceptual expansion}.

%--------------------------------------------------------------------------------
\subsubsection{Co-Creative Generative Systems}
Co-creative generative systems \cite{yannakakis_mixed_2014} let human designers collaborate with procedural generators, enabling designers to guide generation towards desired content. Such systems have been used for generating platformer levels \cite{smith_tanagra_2011}, game maps \cite{liapis_sentient_2013}, \textit{Cut-the-Rope} levels \cite{shaker_ropossum:_2013} and dungeons \cite{baldwin_mixed_2017}. Within PCGML, the Morai Maker \cite{guzdial_general_2017} lets users design Mario levels by collaborating with AI agents based on generative models from past PCGML work. We envision that VAEs can be the basis for similar tools that help designers in making levels that blend properties of multiple games.

\subsubsection{Variational Autoencoders}
Autoencoders \cite{hinton_reducing_2006} are neural nets that learn lower-dimensional representations of data using an unsupervised approach. They consist of an encoder which converts the data into this representation (i.e. the latent space) and a decoder which reconstructs the original data from it. Variational autoencoders (VAEs) \cite{kingma_auto_2013} augment vanilla autoencoders by making the latent space model a probability distribution which allows learning a continuous latent space thus enabling random sampling of outputs as well as interpolation, similar to GANs. Thus a level generation approach as in MarioGAN \cite{volz_evolving_2018} could be implemented using VAEs. Similar to how MarioGAN searched the latent space to find desired levels, the same may be possible with VAEs. Like our approach, \citeauthoryearp{guzdial_explainable_2018} used autoencoders for generation, focusing on generating level structures conforming to specified design patterns. Their work differs in generating Mario-only structures and using a standard autoencoder while we generate segments spanning both Mario and Icarus using a VAE.

\subsubsection{Latent Variable Evolution via CMA-ES}
Latent variable evolution \cite{bontrager_deep_2018} refers to using evolutionary search to find desired vectors in latent space. MarioGAN uses Covariance Matrix Adaptation Evolutionary Strategy (CMA-ES) \cite{hansen_reducing_2003} to find desirable level segments within the latent space of the trained GAN. This allows generating segments that optimize specific features such as Tile Pattern KL-Divergence as introduced by \citeauthoryearp{lucas_tile_2019}. We similarly used CMA-ES to evolve vectors in the latent space of our VAE.

\XTABLEencoding

\XFIGUREall

\section{Approach}

For the remainder of the paper, we refer to \textit{Super Mario Bros.} as SMB and \textit{Kid Icarus} as KI.
%================================================================================
\subsubsection{Dataset}

Level data was taken from the Video Game Level Corpus (VGLC)\footnote{https://github.com/TheVGLC/TheVGLC} \cite{VGLC} and consisted of data from one level (Level 1-1) of \textit{SMB} and and one level (Level 5) of \textit{KI}. We chose the former since it was used in MarioGAN and the latter as it contains all the KI elements which is not true for all KI levels in the corpus. We used only 1 level per game similar to MarioGAN and also due to the exploratory nature of our work, hoping for quick, promising results and leaving more optimal results derived from a larger corpus for future work. Regarding the choice of games, our ultimate goal is blending games from different genres. To do so, it is worth first blending games from the same genre that differ in significant ways. Thus, we use SMB and KI since both are platformers but differ in orientation. Each level in the VGLC is a 2D character array with each tile represented by a specific character. For training, each tile type was encoded first using an integer, as given in Table \ref{XTABLEencoding}, and then using a One-Hot encoding. To create training samples, we used a 16x16 window slid horizontally across the SMB level and vertically across the KI level, thus training the VAE on 16x16 level segments from both games. We chose this window to account for the difference in orientation between the games. Hence, the VAE learned to generate 16x16 level segments rather than entire levels. This may be more conducive to a co-creative approach as it lets designers query the VAE for segments they can assemble as desired rather than the VAE constructing a level with a fixed rule for combining generated segments. In MarioGAN, this is not an issue since all segments are from SMB so a full level can be formed by placing them one after another. It is not obvious how to do so when segments themselves may be a blend of SMB (horizonatal) and KI (vertical) gameplay. Using the sliding window, we obtained 187 SMB and 191 KI training segments, for a total of 378 training segments.

\subsubsection{Training}
The VAE encoder had 2 strided-convolutional layers using batchnorm and LeakyReLU activation. This fed into a 64-dimensional bottleneck (i.e. hidden) layer which fed into the decoder. The decoder had 1 non-strided  followed by 2 strided-convolutional layers and also used batchnorm but used ReLU instead of LeakyReLU. We used the Adam optimizer with a learning rate of 0.001 and binary cross-entropy as the loss function. For evaluation, we compared the VAE with a GAN and a VAE-GAN. The GAN discriminator and generator had similar architectures to the VAE encoder and decoder respectively. The VAE-GAN's encoder and decoder had the same architecture as those of the VAE while its discriminator had the same architecture as that of the GAN. For all models, we used PyTorch \cite{paszke2017automatic} and trained on the 2 levels for 10000 epochs based on MarioGAN using 5000 epochs to train on 1 level.

\XFIGUREprops

\subsubsection{Generation}
The trained VAE generates 16x16 segments within the combined \textit{SMB-KI latent level design space}. Generation involves feeding a 64-dimensional latent vector into the VAE's decoder which outputs a 17x16x16 one-hot encoded array representing the segment. Using argmax along the one-hot encoded dimension gives the 16x16 segment in the integer encoding as in Table \ref{XTABLEencoding}. This can then be converted and stored as an image using tiles from the original games. Thus, like MarioGAN, the VAE can generate segments by using random latent vectors or by using CMA to evolve vectors that optimize given fitness functions, thereby satisfying designer-specified constraints. Unlike GANs, the VAE can also generate segments based on those supplied by designers rather than just using latent vectors. This involves feeding a segment encoded using the VGLC representation into the VAE's encoder which in turn encodes the segment into the learned 64-dimensional latent vector representation. A new segment can then be obtained from this vector using the decoder. Further, one can input two segments, get their corresponding vectors using the encoder and interpolate between them to generate new segments that blend the input segments. Due to these added capabilities, we argue that VAEs are more suited than GANs for co-creative level design systems based on blending different games as it allows designers more explicit control in defining the inputs to the system. Designers may find it more useful to blend or interpolate between segments they define or know the appearance of rather than do so by evolving latent vectors.

\section{Evaluation}
To evaluate our approach, we used the following metrics:  
\begin{itemize}
    \item \textit{Density} - the number of solid tiles in a 16x16 segment. A segment with density of 100\% has all 256 tiles as solid.
    \item \textit{Difficulty} - the number of enemies plus hazards in a 16x16 segment. Based on the dimensions, we defined a segment with 100\% difficulty to have 16 total enemies and hazards.
    \item \textit{Non-Linearity} - measures how well segment topology fits to a line. It is the mean squared error of running linear regression on the highest point of each of the 16 columns of a segment. A zero value indicates perfectly horizontal or linear topology.
    \item \textit{SMB Proportion} - the percentage of non-background SMB tiles in a segment. A segment with 100\% \textit{SMB Proportion} has only SMB tiles while 0\% has only KI tiles. %KI Percentage is 100\% - SMB Percentage.
\end{itemize}
\XFIGUREcorners

These metrics are tile-based properties meant to visualize the generator's expressive range \cite{smith_analyzing_2010} rather than encapsulate formal notions of density, difficulty or non-linearity. Additionally, we compared the VAE's generative performance with that of similar generative models like the GAN and the VAE-GAN. While the GAN cannot encode known levels and segments, it may offer better generation and blending than the VAE in which case using a hybrid model such as the VAE-GAN \cite{larsen_autoencoding_2016} that combines the benefits of both, may be best for blending levels. Thus, we compared these models in terms of their accuracy in evolving desired segments by using CMA-ES to evolve 100 level segments with target values of 0\%, 25\%, 50\%, 75\% and 100\% for each of \textit{Density}, \textit{Difficulty}, \textit{Non-Linearity} and \textit{SMB Proportion} and compared the target values to the actual values of the evolved segments. Further, we compared the models in terms of capturing the latent space spanning both games. For this, we computed the above metrics for segments generated from 10,000 latent vectors drawn uniformly at random from a Gaussian distribution.

\XFIGUREevo

\XFIGUREdensity

\XFIGUREdifficulty

\XFIGUREnonlinearity

\XFIGUREblending

\XFIGUREint
\XFIGUREsint
\XFIGUREmaxtt

\section{Results and Discussion}
Figure \ref{XFIGUREall} depicts the expressive range of the three models in terms of \textit{Density}, \textit{Difficulty}, \textit{Non-Linearity} and \textit{SMB Proportion} in generated segments. The VAE seems to be best at generating segments whose elements are a mix of those from either game while both the GAN and the VAE-GAN generate segments with mostly SMB or mostly KI elements as evidenced by their plots being sparser in the middle than the VAE plot. This is also suggested by Figure \ref{XFIGUREprops}. Approximately 18\% of VAE-generated segments have elements of both games where as this drops to around 8\% for GAN and 5\% for VAE-GAN. This implies that the VAE is better than the other models at capturing the latent space spanning both games as well as the space in between, thus making it the best choice among the three for generating blended segments. We also used corner plots \cite{foreman_corner_2016} to visualize the models as shown in Figure \ref{XFIGUREcorners}. Such plots have been used in past PCGML work \cite{summerville_super_2017} and visualize the output of a model with respect to multiple metrics simultaneously. Based on these, VAE-generated segments adhere more closely to training segments and the space between segments from both games. While the increased variance displayed by segments generated by the GAN and VAE-GAN seems desirable for novelty, as we will discuss, this is due to the segments over-generalizing and ignoring the structures of training segments.

Results of testing the accuracy of evolving segments based on properties are shown in Figure \ref{XFIGUREevo}. Initially, the GAN seems to perform best in terms of \textit{Density}, \textit{Difficulty} and \textit{Non-Linearity}, followed by the VAE and then the VAE-GAN. It is worth noting though that the GAN does better than the VAE only for 100\% \textit{Density} and 75\% and 100\% \textit{Difficulty}. However such values ignore the structures in training levels since actual SMB and KI segments would have neither 100\% solid tiles nor 16 enemies and hazards. This suggests that the VAE's latent space better captures the nature of the training data than the GAN's latent space, thereby struggling to find segments with close to 100\% \textit{Density} while the GAN can do so more easily. It is possible that the GAN wasn't trained enough but since we used the same training data, similar architectures and the same number of epochs for both models, this is another benefit of the VAE as it exhibits better performance with similar training.

In terms of blending desired SMB and KI proportions, none of the models do particularly well in evolving segments that are neither 100\% KI nor 100\% SMB. However, while the VAE does well at least for 50\%, the other two do much worse. These results follow from the discussion on Figure \ref{XFIGUREprops} and suggest that, with similar training, the VAE learns a latent space that is more representative of the game data (based on Figure \ref{XFIGUREevo}) while having more variation to enable better blending (based on Figures \ref{XFIGUREall} and \ref{XFIGUREprops}). Thus, in addition to enabling the encoding of segments, VAEs seem to also be better at generation than GANs in the context of level blending. Example evolved segments for each model for different objectives are shown in Figures \ref{XFIGUREdensity}, \ref{XFIGUREdifficulty}, \ref{XFIGUREnonlinearity} and \ref{XFIGUREblending}. While future work should focus on improving the VAE architecture to better evolve segments spanning all values for \textit{SMB Proportion}, overall, our findings suggest that VAEs are better suited to blending levels from different games than GANs.

\section{Application in Co-Creative Design}
VAEs trained on levels from multiple games could inform co-creative level design systems that let designers make levels by generating and blending level segments representative of all games used for training via the following affordances:

\subsubsection{Interpolation between games}
By encoding level segments into latent vectors, VAEs enable interpolation between vectors representing segments from different games and generation of segments that lie between the latent space of either game, thus having properties of both games i.e. blended segments. For example, one could interpolate between an SMB segment and a KI segment as in Figure \ref{XFIGUREint}.

\subsubsection{Alternative connections between segments}
Interpolating between segments from the same level can generate alternate connections between them. This can help designers edit existing levels by interpolating between two segments from the level to generate new segments that can be combined to form new links between the original two as in Figure \ref{XFIGUREsint}.

\subsubsection{Generating segments satisfying specific properties}
Search within the VAE's latent space can be used to evolve vectors and thus level segments satisfying specific properties. We saw \textit{Difficulty}, \textit{Density} and \textit{Non-Linearity}, but other optimizations are possible such as maximizing the distribution of certain tiles as in Figure \ref{XFIGUREmaxtt}.

\subsubsection{Generating segments with desired proportions of different games}
In addition to optimizing for tile-specific properties, we can also optimize for desired proportions of level elements from each game, as in Figure \ref{XFIGUREblending}.\\

\noindent These affordances make VAEs suited to co-creative level design using latent spaces of multiple games. While GANs can generate segments optimizing specific properties, they do not offer the first two affordances as they only accept latent vectors as input. Also, based on our results, VAEs seem to be better at learning latent spaces spanning multiple games. Moreover, while the LSTM approach can generate entire blended levels, it does not offer any affordance except proportional blending between two games. Even so, this was done by varying the number of segments from each game rather than blending segments themselves. Currently, the VAE is unable to generate whole blended levels. When generating segments blending levels from games with different orientations, it is not obvious how to combine them. The LSTM approach addressed this by training a classifier on SMB and KI level sequences to determine if generated sequences were more SMB-like or KI-like, orienting them accordingly. However, this is harder when segments themselves are blended as in our case. While future work should augment the VAE with entire level generation capabilities, in a co-creative context, segments may offer designers more fine-grained control over level design than whole levels. They are likely to better affect level aesthetics by deciding on the placement of generated segments instead of using entire generated levels.

%================================================================================
\section{Limitations and Future Work}

We trained a VAE on level data from \textit{Super Mario Bros.} and \textit{Kid Icarus} and argued for its use as the basis for a co-creative level design system that enables blending of levels from both games. We discuss limitations and future work below. 

\subsubsection{Playability}
The main limitation of this work is that it ignores playability. This may be less problematic when generating segments rather than whole levels as the designer can place segments such that the level is more playable. However, they should ideally be able to query the generator for segments that maintain playability relative to those already generated. This ties into the issue of blended levels possibly requiring new or blended mechanics to be playable. Investigating methods for blending mechanics is necessary future work and could leverage past work in mechanic generation. \cite{zook_automatic_2014,khalifa_general_2017,smith_variations_2010,guzdial_game_2017}.

\subsubsection{Vector Math in Level Design Space}
Besides interpolation, other vector operations like addition and subtraction can be used in the latent space to generate segments. Such vector arithmetic could enable feature transfer between games by for example adding a vector representing a game 1 feature to a vector representing a game 2 segment. Such interactions would require \textit{disentanglement} i.e. different latent space dimensions encoding different level features. VAEs capable of learning such disentangled representations are called disentangled or $\beta$-VAEs \cite{higgins_beta_2017} and should be explored for this purpose in the future.

\subsubsection{Level Design Tool} In this work, we focused on validating the VAE approach to level blending. The next step is to implement the co-creative level design tool described in the previous section, using the VAE as its foundation.

\subsubsection{Multiple Games and Genres}
Finally, future work could consider blended level design spaces spanning more than two games as well as multiple genres. How would one blend a platformer with an action-adventure game? What games, genres, mechanics and levels exist within the latent space between \textit{Mario} and \textit{Zelda}, for example?

%================================================================================

%================================================================================
\bibliographystyle{aaai}
\bibliography{main}

\end{document}